\documentclass[conference]{IEEEtran}
\ifCLASSINFOpdf
\else
\fi
\usepackage{graphicx} 
\usepackage{multicol}
\usepackage{algorithm}
\usepackage{algpseudocode}
\usepackage{verbatim}
\usepackage{amsmath}
\usepackage{mathtools}
\usepackage{textcomp}
\usepackage{url}
\usepackage{adjustbox}
\usepackage{multirow}

\begin{document}
%
\title{Reuleaux: Robot Base Placement by Reachability Analysis}

\author{\IEEEauthorblockN{Abhijit Makhal*}
\IEEEauthorblockA{Department of Mechanical Engineering\\
Idaho State University\\
Pocatello, Idaho 83209\\
Email: http://makhabhi@isu.edu}
\and
\IEEEauthorblockN{Alex K. Goins}
\IEEEauthorblockA{ROS-Industrial\\
Email: alex.goins@swri.org }
}


%


\maketitle

\begin{abstract}
Before beginning any robot task, users must position the robot's base, a task that now depends entirely on user intuition. While slight perturbation is tolerable for robots with moveable bases, correcting the problem is imperative for fixed-base robots if some essential task sections are out of reach. For mobile manipulation robots, it is necessary to decide on a specific base position before beginning manipulation tasks. 

This paper presents Reuleaux, an open source library for robot reachability analyses and base placement. It reduces the amount of extra repositioning and removes the manual work of identifying potential base locations. Based on the reachability map, base placement locations of a whole robot or only the arm can be efficiently determined. This can be applied to both statically mounted robots, where position of the robot and work piece ensure the maximum amount of work performed, and to mobile robots, where the maximum amount of workable area can be reached.  Solutions are not limited only to vertically constrained placement, since complicated robotics tasks require the base to be placed at unique poses based on task demand. 

All Reuleaux library methods were tested on different robots of different specifications and evaluated for tasks in simulation and real world environment. Evaluation results indicate that Reuleaux had significantly improved performance than prior existing methods in terms of time-efficiency and range of applicability. 
\end{abstract}


%
\IEEEpeerreviewmaketitle

\section{Introduction}
The robotic industry's ongoing advancements depends upon state-of-the art equipments- with unique specifications and capabilities. As the hardware of the robotic systems are developed, so as the softwares are metamorphosed. From the perspective of an industry or household user, however, the main goal of deploying a robotic system remains the same: to successfully perform the desired task. Tasks can range from intense industrial work such as welding, packaging, and manipulation to relatively smaller scale of picking, placing or grasping. For users to accept and employ a robotic system, it must, (1) precisely execute the task, and (2) integrate well into the user's workspace. For both features, the user and the robot should be aware of information about the robot's reach and the workspace. Without these knowledge, deployment of the robot depends solely on user intuition. An incorrect intuition can lead to human casualty or catastrophic damage to the robotic system and workplace.

The position of mobile robotic system with wheels or legs, can be changed, so the system is already facilitated with capabilities of moving itself. 
On the other hand, a system with a fixed base or a system adamant to move, the task of shifting is completely relied on the user, which can be highly intimidating if the task cannot be performed from a specified location.

Non-expert users who do not have an in-depth understanding of robot kinematics and the reachability challenges of robots, might hold misconceptions that the workspace of a robotic manipulator is sphere-shaped when the radius of the arm is fully extended and in this region the robot's end-effector can move freely. On the contrary, the robotic arm's workspace depends fully on the constraints posed on its arm joints. In the workspace, there are few positions where the arm can reach freely; in some sections of the workspace, the arm faces singularity. In this paper, the authors conferred few methods by which the robot's workspace could be fully utilized and a suitable placement of the robot or manipulator base can be achieved.

The paper makes the following three contributions to the field of robotic workspace analysis and base placement, 1) A method to generate and analyze the precise reachability of any generic robotic arm in a time-efficient manner, 2) A method to localize the feasible base positions of the whole robot or its manipulator for any given user task and  3) A comparison and evaluation of both methods on different robots with distinct characteristics.

\begin{figure*}
\centering
\includegraphics[width=\textwidth,height=6cm]{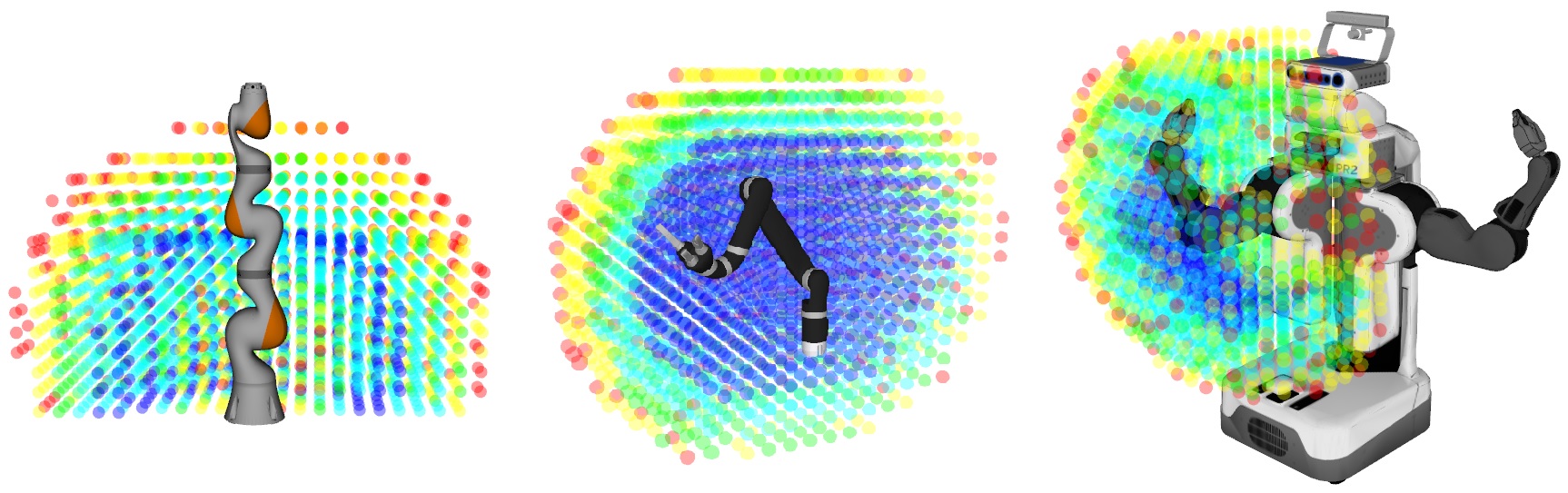}
\caption{Reachability Map representation. From left to right a) LWR 7DOF arm with self-collision checking, b) 6DOF JACO arm without self-collision checking, c) right arm (7DOF) of PR2 mobile manipulator with self-collision checking. (Color representation with increasing reachability- Red, Yellow, Green, SkyBlue and Blue)}
\label{fig:reach_map}
\end{figure*}

The rest of the paper is organized as follows: Section \ref{Related_work}
presents relevant related work for robot workspace analysis and base
placement. Section \ref{reachability_map} explains the methodology incorporated in generating reachability map and Section \ref{inv_reachability_map} discusses
the procedure for constructing an inverse reachability map. The
methods and procedures for finding the optimal base position
are presented in Section \ref{base_placement}, while results of the reachability
analysis and base placement methods as well as a comparison of results are documented in Section \ref{results}. We conclude with Section \ref{conclusions}'s  discussion
of future work and work bottlenecks.

\section{Related work}  \label{Related_work}

One of the challenges in developing a useful robotic system is designing a platform or workspace that lets the robot to effectively complete the desired task. 
Initial work on
path planning and reachability was typically performed for
simple grasp points on objects \cite{6651543}, which involves creating a
map representing the areas of high dexterity for the manipulators. This work was then extended to the use of reachability maps to solve the inverse reachability task \cite{7139993} and \cite{diankov_thesis}, where the optimal base placement was found in order to perform the desired grasp on an object. Work done by \cite{4399105} examined ways to simplify the reachability map by generating a capability map, a simplified structure that permits faster and more efficient searches of the map in order to
solve the inverse reachability problem. 

Though these methods are limited. They solve only the general problem of whether or not an inverse kinematics solution was found. A given grasp location could mean that some or all of the solutions found could be non-optimal (near singularities or joint limits). The work of \cite{Yoshikawa:1990:FRA:96732} presents the concept of manipulability ellipsoid measure, which can be derived from a Jacobian matrix of manipulators. The size of the ellipsoid and the principal axes represent the manipulation ability in a certain configuration and are thus used to determine the effectiveness of a grasp at a given location. This work was extended in terms of grasping and manipulation
in \cite{Vahrenkamp:2015:RRW:2720528.2720571}.

To overcome the single grasp location issue, several papers extended the inverse reachability problem to solve for trajectories. Various methods have been used to search the reachability map in order to find the location where the desired trajectory can be executed with the highest level of dexterity. \cite{Vahrenkamp2013RobotPB} uses sampling of the trajectory to find and overlay multiple base placements; \cite{4650617} uses a
pattern search to fit the trajectory into the area representing the field of high dexterity; and \cite{5379601} uses a cross-correlation technique to fit the desired trajectory to the model of the robot reachability map. Further improvements to the reachability map method were developed by \cite{7353564} to include the
ability to add a transform offset from the original end-effector
location, which is useful if the robot is grasping a tool
with a non-zero length. 

These methods allow planning for simple tasks, but they require a task to be completely specified prior to evaluation. \cite{Porges2015ReachabilityAD} explores the case where a specific trajectory is not given; rather the task has been simplified into a generalized workspace environment where the robot must work. Here, competing constraints are given.
However the operator still evaluates and confirms the final base placement of the robot on the mobile platform. For a simple operation with few operating points, this can
be done manually, but for more complex parts and tasks such as welding pipes or assembling parts in constrained spaces, a manual approach for validation may not be feasible.

\section{REACHABILITY MAP} \label{reachability_map}
A reachability map is a collection of all poses that the robot's end effector can reach. To accommodate the infinite number of reachable poses, similar poses are clustered into a sphere structure suitable for visualization and easy to access. The structure also captures the directional information of reachable poses, because several positions can be reached only from a certain direction. An individual sphere is represented as a multimap data structure, which holds information about the sphere's position, all reachable poses belonging to the sphere, and the reachability measure of that sphere. 

To create a reachability map, a method similar to the work presented in \cite{4399105} is incorporated where the robot's workspace is discretized, sampled for reachable poses and information of reachable poses are stored in a structured manner. The reachability map structure of three different robots created with the Reuleaux library is shown in Fig.\ref{fig:reach_map} 

\begin{algorithm}  
\caption{Generate Reachability Map} input (URDF of robot)
\label{reach_map_algorithm}
\begin{algorithmic}[1]
\Procedure{$GenerateReachM ap$} {}
\\create VoxelStructure $V$
\For{each voxel $v_i$ in $V$ }
\State create sphere $s$\
\State checkforSelf Collision($s$)\
\State Store $s_{filtered}$ in $S$
\EndFor
\For {each $s_i$ in $S$ }
\State Sample surface and generate poses $P$
\For {each $j$ in $P$ }
\State findIKSolution ($P_j$)
\If {solution }
\State Store ($s_i$, $P_j$)
\EndIf
\State $d_i$ = $FindReachabilityMeasure$($s_i$)
\State Store $d_i$ with ($s_i$, $P_j$)  in map
\EndFor
\EndFor
\EndProcedure
\end{algorithmic}
\end{algorithm}

\subsection{Workspace Voxelization}
Input to the creation of a robot reachability map procedure is the Unified Robot Description format (URDF) model of the robot, from which a detailed robot description is obtained. The robot's hypothesized workspace is first voxelized to create a square structure around the robot. The voxelization process is performed by octree \cite{hornung13auro}, a hierarchical data structure enabling spatial partitioning and searching between voxels. The root node of the octree is at the base of the manipulator, and every node is connected to its eight children. The tree is extended to the overestimation of the diameter of the robot arm in an extended state. The size of the voxels required are task dependent and thus user defined. Tasks requiring a high degree of accuracy will need a smaller voxel size in order to provide more accurate results in the final base placement location. The centers of voxels are determined, and a sphere with a radius of the voxel's resolution is placed in every voxel.

\subsection{Self Collision Checking}
All the spheres collected by the voxelization procedure do not indicate reachable workspace to be considered. The workspace regions where the robot body is present should not be included in the workspace because end-effectors cannot reach those sections due to collisions. For filtering out such sections, as a preprocessing step, the robot body is modelled as a collection of triangular meshes and checked for collisions with the voxelized sphere center. The sphere centers in collisions with robot body (excluding the arm considered for reachability analysis) are opted out of the workspace structure; they are not further discretized for reachability analysis. 

A collision checking library FCL \cite{FCL} is used for fast collision checking. This method provides an advantage in terms of time efficiency since most other methods such as \cite{4399105} and \cite{diankov_thesis}, check for collision only when obtaining solutions from poses. This step drastically reduces the number of spheres to be searched. 

In our method, self-collision is also included while checking for reachability of the poses. Certain poses may not be inside the robot body, but to reach the pose, some part of the robot may be in collision. Such poses can be considered unreachable. It is worth noticing in Fig.\ref{fig:reach_map} that reachability maps created without self-collision represent overall symmetric structures that consider higher reachability at the robot's base, e.g. the reachability map of JACO arm (Fig.\ref{fig:reach_map}b). Due to joint limits and self-collision, the area of the robot base should be less reachable, which can be seen in Fig.\ref{fig:reach_map}a, the reachability map of the LWR arm. The reachability map of the PR2 robot in Fig.\ref{fig:reach_map}c shows an inconsistent structure since the sphere centers within the robot's body are filtered out in this step.

\subsection{Workspace Sampling}

One of the most popular ways to determine robot's possible reach position is by forward kinematics (FK), where the joints of the manipulator can be uniformly sampled
and the configuration of the tool center point (TCP) can be stored in an efficient structure to represent the reachability workspace. The number of TCP positions in a voxel gathered from FK  was considered to be a suitable representation of efficient reachability in certain voxelized sections. In \cite{4399105}, it is proven invalid for most cases: If the positions in the voxel represents a singular configuration, any large step in the configuration space would cause small steps in the Cartesian space, leading to abundance of poses in a single region. Also, uniform sampling of the joint space does not
guarantee uniform sampling in the Cartesian space.

\begin{figure}[h!]
\centering
\includegraphics[scale=0.23]{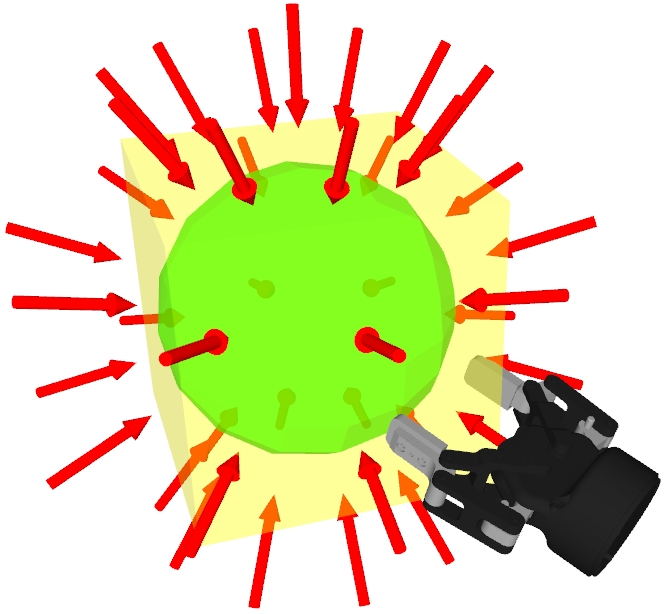}
\caption{Discretization of a single voxel in the workspace. The cube(yellow) is the voxel to be searched. A sphere (green) is fitted inside the cube. The sphere is discretized with probable poses (red arrow) }
\label{fig:discretization}
\end{figure}

For uniformly sampling the task space, the spheres representing the voxels are sampled for uniform point distribution on the sphere by the method presented in \cite{sp_distribution}. From every point of the distribution, a direction is assumed towards the center of the sphere. As depicted in Fig.\ref{fig:discretization}, an individual voxel of the discretized workspace is fitted with a sphere, and the sphere is sampled for points on its surface. From an individual point, a direction is considered towards the center of the sphere represented by a red arrow. Considering a frame, the direction towards the center of the sphere is the $z$-axis, while the $x$ and $y$ axes are tangential to the sphere surface. 

All the frames created in the previous step are collected and searched for inverse kinematics(IK) solutions. A closed-form analytical inverse kinematics solver IKFast\cite{diankov_thesis} is employed to find inverse kinematics solutions. For manipulators where analytical solutions are not available, a numerical solver KDL\cite{KDL} is used which is less time-efficient due of the nature of the solver. All reachable poses that return a valid joint solution are stored in a multimap datastructure with their corresponding sphere centers.

\subsection{Measure of Reachability}
The reachability map generation procedure is simplified in Algorithm \ref{reach_map_algorithm}. 
A reachability metric is been proposed for a given voxel based on the fraction of discrete poses in the sphere centered at that voxel that are reachable to the robot. In equation(\ref{eq:reach_measure}), the reachability measure is termed $D$, where the number of discretized frames of a sphere is represented by $N$ and the number of reachable frames calculated is $R$.
\begin{equation}
\label{eq:reach_measure}
D = (R/N) * 100
\end{equation}

Based on the calculated $D$, spheres are assigned to different colors ranging from red to blue, where blue represents higher reachability(i.e., the spheres with most reachable poses) and red represents lower reachability. In terms of decreasing reachability, the colors are in the sequence of blue, skyblue, green, yellow and red. 

\section{INVERSE REACHABILITY MAP(IRMs)} \label{inv_reachability_map}

Considering every pose in the reachability map to be a transformation matrix from the origin of the robot, inverse transformation on every pose are performed to create an inverse reachability map (IRM). The IRM is represented using the same data structure as the reachability map, consisting of poses and spheres. The 6D pose of the TCP is represented as:

\begin{equation}
T_{global}^{TCP} = \Big\{x,y,z,\rho, \phi, \theta \Big\}
\end{equation}

\begin{equation}
T^{-1} = T_{TCP}^{global} 
\end{equation}

where $\{ x,y,z $\} is the pose of the end effector and $\{\rho, \phi, \theta $\} is the orientation represented in Euler angles. In the IRM, $T^{-1}$ is stored which will be later used in the base placement scenario where task poses are defined by the user. To create a sphere structure a reachability measurement similar to the reachability map, we consider every transformed frame to be a position $\{ x,y,z $\} in the environment. 

\section{BASE PLACEMENT} \label{base_placement}
In this section, potential base poses for a given is be searched, where the task poses are completely dependent on the user. Task poses can be a discretized trajectory or can be an indication of certain regions the robot has to reach for pick and place task. The inverse reachability consisting of all the transformations is the main ingredient of the base placement process. Task poses $task_i$ are transformed using transformations $T^{-1}_j$ of the inverse reachability map to create an union map of all the potential base poses $B_{ij}$ .

\begin{equation}
B_{ij} = task_i * T_j ^ {-1}
\end{equation}

All the poses $B_{ij}$ are considered to be points and stored in an octree structure; they are processed through a nearest neighbour (NN) algorithm that clusters together all poses in the same voxel. This procedure is the inverse process of sphere discretization process. In the previous process, poses are obtained from spheres; here the spheres are acquired from transformed poses. The color of the voxels is determined by placebase index in eq(\ref{eq:base_reach_cal}).

 The colors are also arranged in the same sequence as in the reachability map. However, instead of representing reachability, here the different colors represent the probability of suitable base placements from where the task poses can be reached. The procedure for creating union map is explained in algorithm.\ref{union_map_algorithm}
 
\begin{equation}
\label{eq:base_reach_cal}
d = \begin{cases} \displaystyle \frac{n*b_{max}}{b_{max}-b_{min}}*100,& \text{if  } d \geq 1\\
     0, & \text{otherwise}
     \end{cases}
\end{equation}

where $b_{max}$ is the maximum number of possible base poses in a sphere, $b_{min}$ is minimum number of possible base poses in a sphere, and $n$ is the number of base poses in the sphere. The size of the union map is the size of the inverse reachability map times the number of task poses. 

\begin{figure}[h!]
\centering
\includegraphics[scale=0.2]{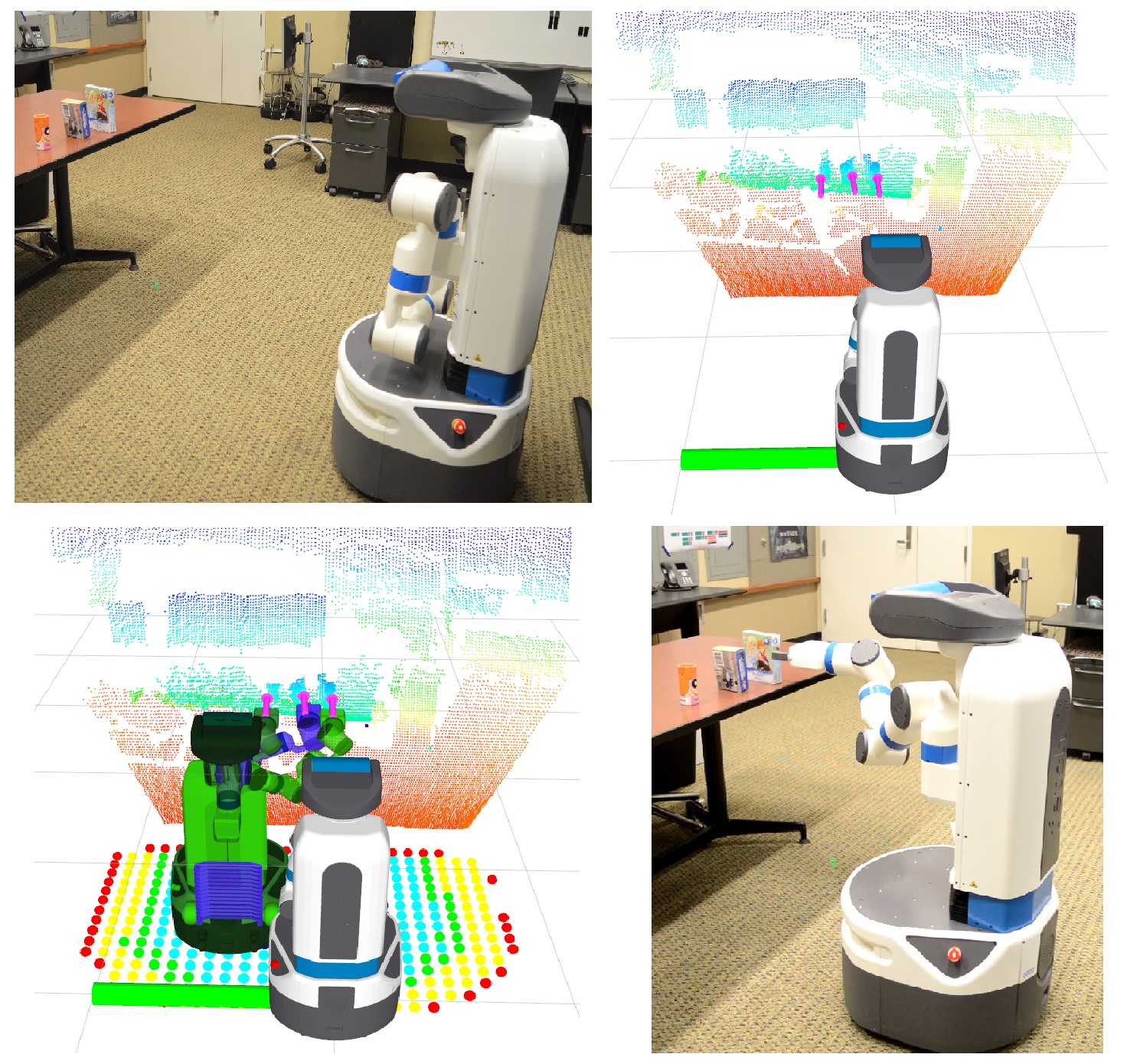}
\caption{Task with real Fetch robot. From left to right, top to bottom. a)Real Fetch robot with 3 objects outside of reachable workspace, b)The visualization of the environment with arrows pointing to grasp poses, c) The inverse reachability map with robot models indicating solution feasibility for all task poses in rviz, d) The robot is reaching for one of the objects after base placement}
\label{fig:fetch_task}
\end{figure}

\begin{algorithm}  
\caption{Union Map Construction} input ($IRM$ , task poses $task_i$)
\label{union_map_algorithm}
\begin{algorithmic}[1]
\Procedure{$createUnionMap$} {}
\For{each task pose $task_i$ }
\State Transform IRM by $task_i$ and obtain $b_{ij}$\
\State Cluster poses $b_{ij}$ by NN and assign to spheres
\State Calculate PlaceBase index for every sphere $S_k$
\State Update in Union map $\langle S_k, b_{ij}, d\rangle$
\EndFor
\EndProcedure
\end{algorithmic}
\end{algorithm}

As explained in Algorithm \ref{union_map_algorithm}, to search through all the spheres and poses for finding probable base locations for the specified task, multiple methods are incorporated. For every method, $task_i$ $\in$ $TASK$ are task poses, $b_j$ $\in$ $B$ are base poses and $s_k$ $\in$  $S$ are spheres. $n$ is the desired number of base locations. For ease of calculation, another parameter, $m$, is also defined, which sets the number of spheres to be searched with the highest placebase index score; when task poses increase, the number of spheres also increases. It is important to note that $n$ should always be lower than $m$.

\begin{figure}[h!]
\centering
\includegraphics[scale=0.2]{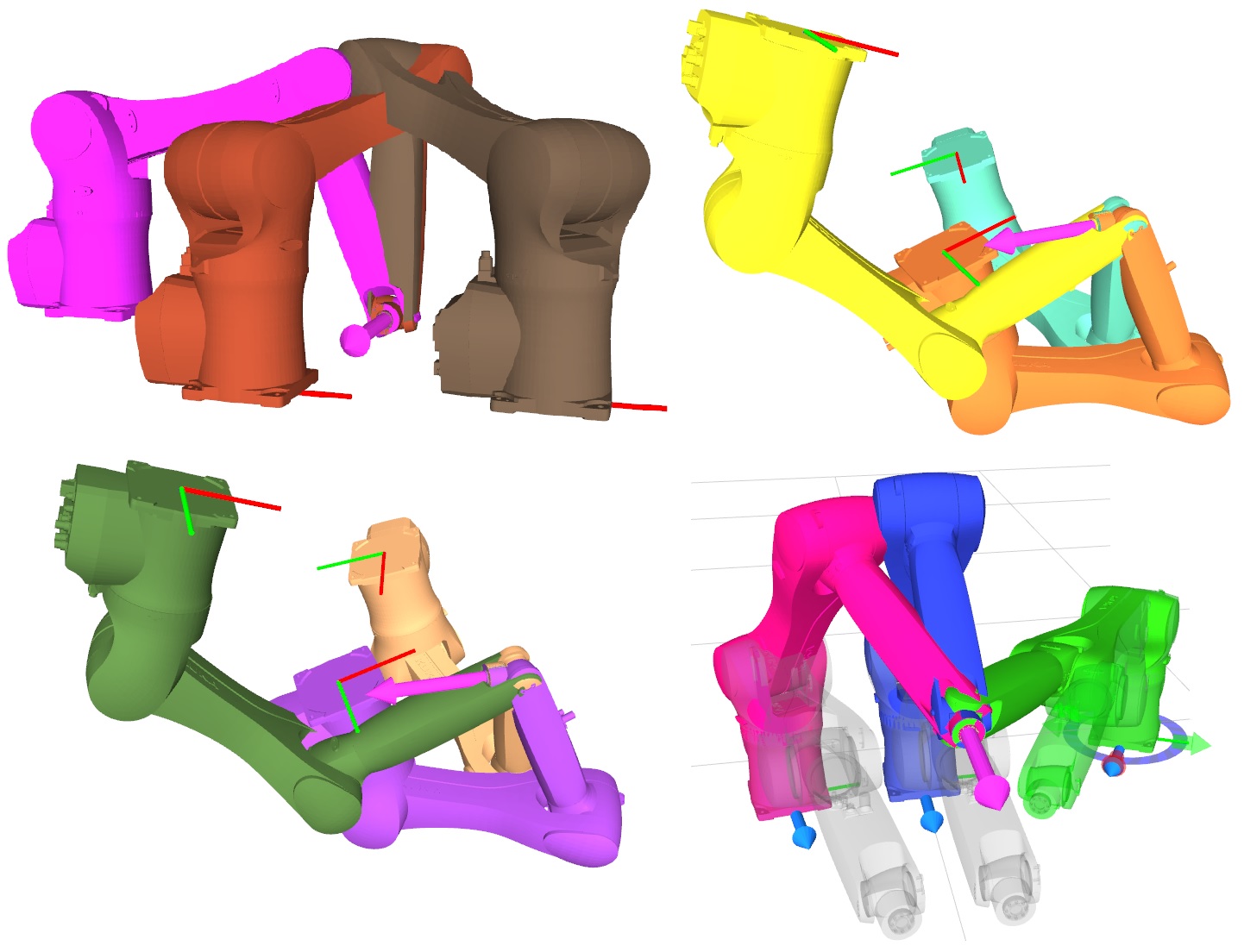}
\caption{For a single task pose (magenta arrow), 3 floating base locations have been searched for the Kuka kr10 arm a) by PCA method b) GraspReachability method, c) IkSolution method, d) user intuition method screen by human input. The manipulator base poses are shown with axes. $z$ axis is inside the robot model }
\label{fig:4_methods}
\end{figure}

\subsection{Placing Floating Base}
In the $Principal Component Analysis(PCA)$ method, all $m$ numbers of spheres with highest placebase index are searched for probable base locations. From $m$ spheres, the orientations $\{\rho_m, \phi_m, \theta_m$\} of base locations are calculated using PCA method on each calculated spheres, $s$, while the positions $\{x_m, y_m, z_m$\} are considered to be sphere centers. The dominant eigen vectors represent the axis of rotation by the which the base position is transformed from the origin. By normalizing the orientations, base locations from the PCA method are set as vertically parallel to the ground while fixing the rotations in the $x$ and $y$ axis. The distances of the base locations varies from the $xy$ plane(the floor) and the only rotation searched for is in the $z$ axis. For every constructed probable base pose $b_m$, all $task_i$ are searched for IK solutions, and the number of found solutions for each $task_i$ stored as a cumulative score. The $n$ poses with the highest scores are considered to be probable base locations. The $GraspReachabilityScore$ method is based on the simplistic assumption that from all calculated optimal base locations, the robot should reach all task poses. Instead of constructing new poses, all the  poses in $m$ spheres are searched for their reach towards all $task_i$. The number of successful reaches is counted and stored. The $n$ poses with the highest scores are considered to be final base locations. Instead of counting the highest number of reaches, in $IkSolutonScore$ method, the total number of IK solutions is considered as score, since the incorporated IK searching method provides maximum of 8 solutions. If there is such a task pose, which is almost non-reachable, the number of solutions decreases with unreachability. The $n$ number of base poses with the highest scores are considered to be the final base locations.

\begin{figure}[h!]
\centering
\includegraphics[scale=0.23]{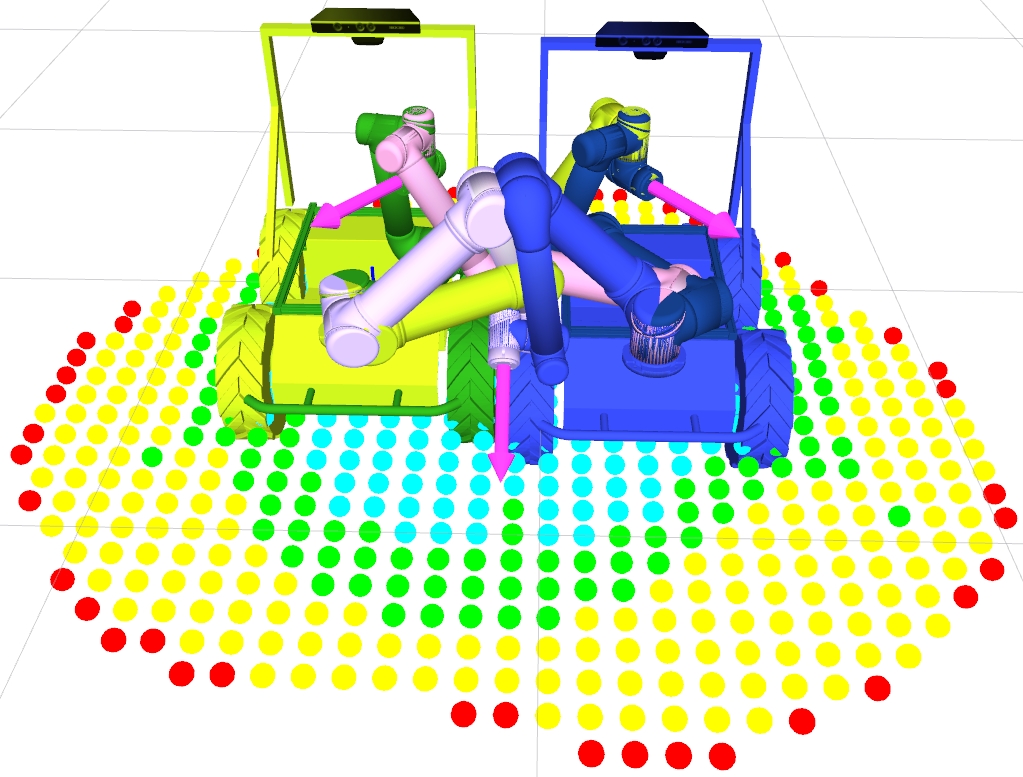}
\caption{The husky robot with an UR5 arm. Two base locations are searched for 3 task poses(magenta arrow). The manipulator configurations from the the left robot model(green) are :a)Dark green, b)green c)white and from the right robot model(blue) are:1)Dark blue, 2)blue 3)pink}
\label{fig:husky2}
\end{figure}

\begin{figure*}
\centering
\includegraphics[width=\textwidth,height=5cm]{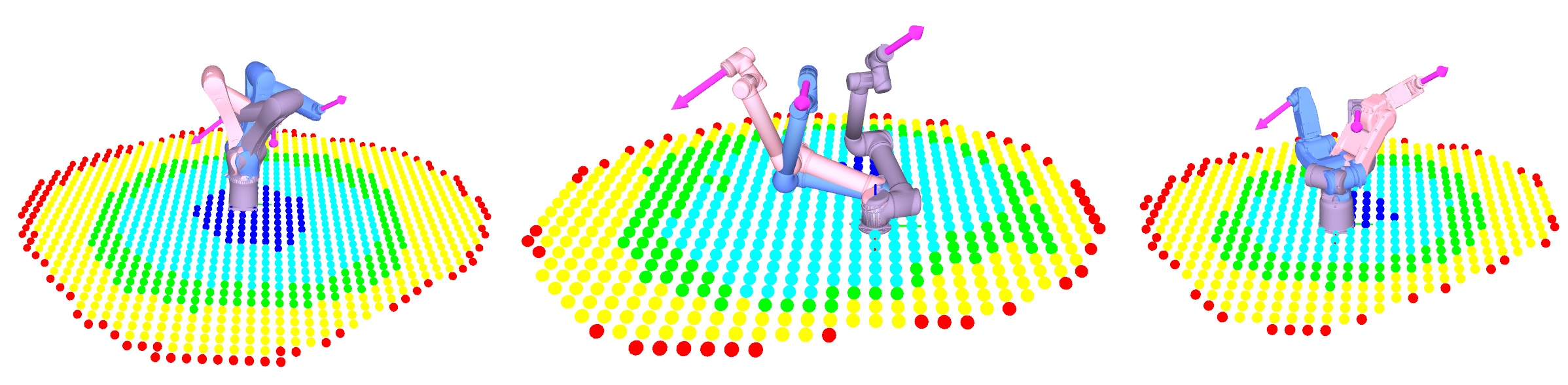}
\caption{Base Placement task on 3 different robots. The task is represented by 3 poses in a table environment. The optimal base placement is shown on a) Kuka
KR6, b)Universal UR5 and c)Motoman mh5}
\label{fig:4_diff_robot}
\end{figure*}

\subsection{Placing Base on the Ground}
Instead of finding a suitable place location for the robot in variable $z$ locations, the $verticalRobotModel$ approach locates probable base locations for the robot on the ground. To solve for robot base location solutions, the transformation from arm base to robot base should be known.

\begin{equation}
\label{T_final}
B_{ij} = task_i * (T_{global} ^ {TCP})^{-1} * (T_{robotbase}^{armbase})^{-1}
\end{equation}

Instead of creating a union map with respect to manipulator's base, the poses at the IRM are also transformed with another extra step by $(T_{robotbase}^{armbase})^{-1}$ to consider at the base location of the full robot shown in eq(\ref{T_final}).
The union map can then be constructed at the robot base location, and the  placebase index is calculated for all $s$ in $S$. As the base of most of the wheeled or legged robot models maintain a parallel stance with the ground, the union map can be transferred from movement in 6D Cartesian space SE(3) to 3DOF space SO(2),where the only movement allowed is the rotation about the $z$-axis and translation in $x$ or $y$-axes. The union map is sliced through a horizontal axis w.r.t to ground. The 3D union map now can be considered as a 2D union map.

\begin{algorithm}  
\caption{Optimal Robot base by VerticalRobotModel
method} input (Union Map, task poses $task_i$, desired number for base locations $n$, spheres to be searched $m$,robot model) output($n$ base locations)
\label{verticalRobot_algorithm}
\begin{algorithmic}[1]
\Procedure{$findbasebyVerticalRobotModel$} {}
\If {$n$ \textgreater{} 0 and $task_i$ \textgreater{} 0}
\State find $T$ between arm base and robot base\
\State Transform IRM by $T$ and $task_i$
\State Slice union map by 2D
\State find $s_m$ with Maximum PlaceBaseIndex
\EndIf
\For{each $k$ in $S_m$} 
\State Uniform sample rotations in $z$ axis and obtain $b_j$
\For{each $b_j$ in $k$} 
\State Calculate score of reachability \
\EndFor
\EndFor
\EndProcedure
\end{algorithmic}
\end{algorithm}

Despite the dimensionality reduction process, the 2D union map still contains the information of placebase index. The $m$ highest placebase indexed spheres are considered to be the optimal base placement locations. As the orientation is constrained in only one direction, the orientation direction is sampled though an uniform interval to generate multiple base poses on a single sphere. From all calculated robot base positions, the reachability of the task poses are considered and the poses unable to reach all task poses are discarded. From the highest scored poses, $n$ poses are selected as final base pose.


A typical condition is presented in Fig.\ref{fig:husky2}, where 2 base locations are searched with 3 task poses. Here, the robot has a Husky robot as a base, and a UR5 arm is the manipulator. To consider optimal base locations, there should be 6 manipulator states present for 3 task poses and 2 base locations . The robot solutions for the task poses are a) left: Dark green and pink, b) middle: Blue and white, c) right: Dark blue and green.

\section{Results} \label{results}
The Reuleaux map generation library was applied on several robots with different specifications and sizes. Without the collision checking facility, the average reachability map generation time was around 156 sec. However, without self collision, the generated map was just a collection of reachable poses, which is disadvantageous. Since there are no standard metric to measure the efficiency of a reachability map, we can consider time efficiency and generalizability as the metrics. As the reachability map can be generated offline, its utility lies in using the map in other tasks.

\begin{table}[h!]
\caption{REACHABILITY MAP GENERATION}
\begin{center}
\begin{footnotesize}
\begin{tabular}{ |c||c|c|c|c| }
\hline
Method & Robot & (x100000) & (x1000)  & Time(min)\\ 
        &       & Poses  & spheres  &   \\
         &       & processed & created & \\
\hline
Reuleaux & PR2 & 20.938 & 2552 & 124.31 \\ 
          & LWR & 13.718 & 5127 & 160.41 \\
           & UR5 & 7.636 & 5017 & 143.27 \\  
\hline
Diankov et.al[4] & PR2 & 205.48 & - & 490.07 \\ 
                  & LWR & 145.727 & - &  427.11\\
                   & UR5 & 123.513 & - & 371.38 \\  
                  
\hline
Zacharias et al[3] & PR2 & 104.69 & 2680 & 405.47 \\ 
                    & LWR & 68.59 & 5213 & 542.18 \\ 
                     & UR5 & 38.18 & 4883 & 413.23 \\ 
\hline
\end{tabular}
\label{tab:ReachMap}
\end{footnotesize}
\end{center}
\end{table}

 In Table \ref{tab:ReachMap} we represent an analysis of Reuleaux's reachability map generation compared to two methods presented in \cite{diankov_thesis} and \cite{4399105} on three different robots. PR2 right arm is a 7DOF manipulator, rigidly connected to the body. The LWR and UR5 are considered on the ground. While the DOF of LRW is 7, the UR5 has only 6ODF. In method \cite{diankov_thesis}, available as open-source library, reachability is not represented as sphere; instead it is represented as area. The opportunity to set the desired options for creating maps is only limited to saving joint solutions and setting a maximum radius. On the other hand, \cite{4399105} represents reachability as a sphere and in extension with different shape representations. Because this method is most similar to our approach, we also represent their work as spheres.
 
In Reuleux, the creation of a reachability map is based on the desired resolution of the voxels and maximum radius. Resolution is a very useful parameter as it determines the size of the voxel, which later creates the size of the spheres. For all our experiments, we set the resolution of voxels at $0.08m$ and maximum radius at $1m$. For convenience, our implementation of \cite{4399105} also uses the same resolution. For all the experments, the IKFast library was used to obtain ikfast solutions.

In Table \ref{tab:ReachMap}, the significant difference in the number of poses processed stems from fact that \cite{diankov_thesis} obtained the poses by default parameters and in \cite{4399105} all poses are rotated by 5 degrees in $z$-direction to obtain extra poses. Regarding the time difference, in our system we have filtered out the spheres on the robot body in a preprocessing step.

\begin{table}[h!]
\caption{MANIPULATOR BASE PLACEMENT PERFORMANCE}
\begin{center}
\begin{footnotesize}
\begin{tabular}{ |c||c|c|c|c| }
\hline
Method & Task Poses & Reachable & score & Time(sec)\\ 
       &            & Poses & & \\

\hline
PCA & 2 & 10/10 & 97.28 & 1.45 \\ 
    & 4 & 20/20 & 93.45 & 2.11 \\ 
\hline
GraspReachability & 2 & 9/10 & 89.7 & 1.77 \\ 
                  & 4 & 17/20 & 85.1 & 2.64 \\ 
                  
\hline
IKsolution & 2 & 10/10 & 96.52 & 1.82 \\ 
           & 4 & 19/20 & 96.27 & 2.79 \\ 
\hline

\end{tabular}
\label{tab:manipPerformance}
\end{footnotesize}
\end{center}
\end{table}

\begin{table*}[ht]
\caption{ROBOT BASE PLACEMENT PERFORMANCE}
\begin{footnotesize}
\centering
        \begin{tabular}{|c||c|c|c|c|c|}
            \hline
             \multirow{2}{*}{robot}& \multirow{2}{*}{reachable}& \multicolumn{4}{c}{Time(sec)} \vline  \\\cline{3-6} 
                                   & {task poses}& base calculation & soln validation & Reach base & Reach task\\
            \hline
                        PR2 (sim)& 4/6 & 21.8s & 0.4s & 7.4s & 7.1s \\\cline{2-6} 
                                 & 6/6 & 18.2s & 0.7s & 5.2s & 6.23s \\\cline{2-6}
                                 & 6/6 & 19.1s & 0.52s & 3.1s & 2.1s \\\cline{2-6}
                                 & 5/6 & 18.6s & 0.7s & 6.5s & 4.78s\\
                                 \hline
                                 \hline
                     Fetch (real)& 3/3 & 9.23s & 0.1s & 4.12 & 6.6s\\\cline{2-6} 
                                 & 2/3 & 8.71s & 0.2s & 3.2s & 7.87s\\\cline{2-6} 
                                 & 1/3 & 8.23s & 0.11s & 7.23s & 9.23s\\\cline{2-6} 
                                 & 3/3 & 8.6s & 0.13s & 4.53s & 7.21s\\\cline{2-6} 
                                 \hline
                                
        \end{tabular}
   \label{tab:real_exp}
 
    \end{footnotesize}
\end{table*}

\begin{figure}[h!]
\centering
\includegraphics[scale=0.19]{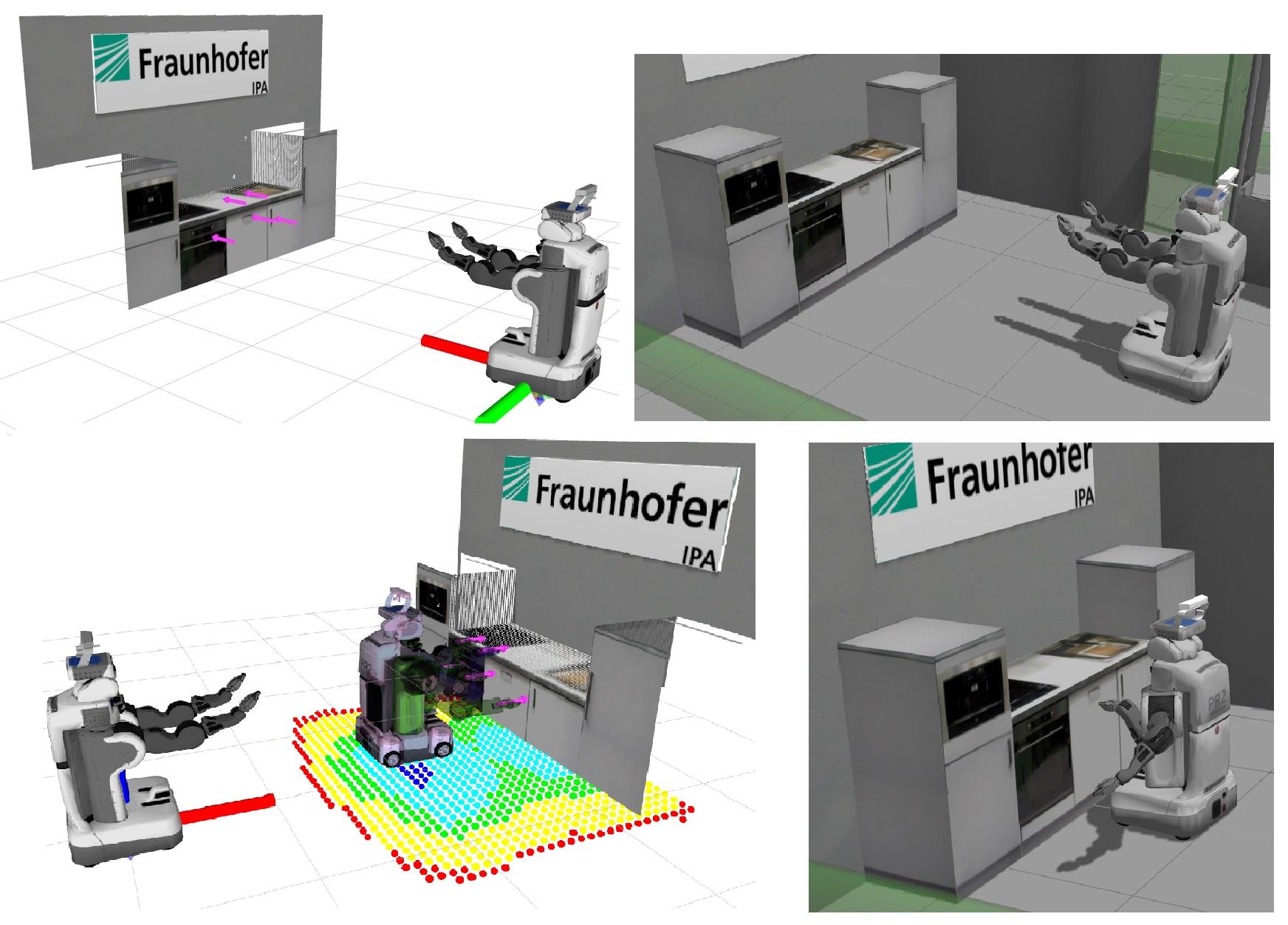}
\caption{Base Placement in simulation. From left to right, top to bottom. a) The rviz view of the environment with arrows pointing to task  poses, b) Simulated PR2 robot in a kitchen environment outside of reachable workspace, c) Visualization of the inverse reachability map with robot models indicating solution feasibility for all task poses, d) The robot is reaching for one of the task pose after base placement}
\label{fig:pr2}
\end{figure}

For all base placement experiments, task poses are decided based on reach tasks. In the simulation scenario, an individual task pose represents a region the robot must access. To represent a motion plan in terms of poses, the trajectory must be uniformly sampled, which is beyond the scope of this paper. For example, in the simulation scenario, shown in Fig.\ref{fig:pr2}, the tasks with magenta arrows represent different sections of the kitchen, such as the sink, oven and drawer. At the initial condition, the tasks were out of robot's reachable workspace. The intent of the system is to find the optimal base position from which all the task poses could be reached. To insert the task poses in the environment, we utilize the depth camera situated on the robot. If a depth camera is not present in the environment, an additional depth camera could be added to the simulation environment, and task poses could be inserted based on the point cloud view from the additional camera.

In Table \ref{tab:manipPerformance}, the base placement task on a table environment is performed by all three floating base placement methods. The result suggests that PCA method is best suited for placing only manipulators in different $z$ locations. So, if user has the opportunity to change the manipulator base height w.r.t to the ground, the PCA method can find the probable placement location for any given task. 
The solutions from the GraspReachabilityScore and IkSolutionScore methods can not be considered as optimal, as they are in turn inverse transformation of the task poses. So including redundancy of a manipulator, the solutions can be infinite. Though from the solutions of this methods, all the task poses are still reachable. 

\begin{table}[h!]
\caption{ROBOT BASE PLACEMENT PERFORMANCE}
\begin{center}
\begin{footnotesize}
\begin{tabular}{ |c||c|c|c|c| }
\hline
Method & Task Poses & Reachable & score & Time(sec)\\ 
       &            & Solutions  & & \\

\hline
Reuleaux & 2 & 10/10 & 97.28 & 1.45 \\ 
    & 4 & 20/20 & 93.45 & 2.11 \\ 
    
\hline
Vahrenkamp & 2 & 9/10 & 89.7 & 1.77 \\ 
              et al[7]    & 4 & 17/20 & 81.17 & 2.64 \\ 
                  
\hline
User1 & 2 & 10/10 & 96.52 & 1.82 \\ 
           & 4 & 19/20 & 95.36 & 2.79 \\

\hline
User1 & 2 & 9/10 & 92.8 & 1.75 \\ 
           & 4 & 19/20 & 94.22 & 3.20 \\

\hline
User1 & 2 & 10/10 & 97.4 & 1.84 \\ 
           & 4 & 18/20 & 79.23 & 3.11 \\ 
\hline

\end{tabular}
\label{tab:basePlacePerformance}
\end{footnotesize}
\end{center}
\end{table}

To evaluate the robot base placement method, we compared our results with methods presented in \cite{Vahrenkamp2013RobotPB}, with a table environment where the robot has to reach different section of the table. For consistency, all task poses are placed on the side of the table and every task pose was set to a different orientation. Fig.\ref{fig:4_diff_robot} shows the task poses for this task on different robots. To prove effectiveness, we employed a new method $userIntutition$ and provided a simple userinterface where the user can drag multiple robot models to set the optimal base placement for a task. We evaluated results from 3 different human users by userIntuition method and their average scores based on 5 tries are also presented in table \ref{tab:basePlacePerformance}. The reader is encouraged to evaluate the $userIntutition$ method from the official repository for the base placement task. The interface for the userIntutition method has been shown in Fig.\ref{fig:4_methods}d. The scores are calculated by IkSolutonScore methods.

We also validated our base placement method: (1) in simulations on a PR2 robot in Franhofer IPA Kitchen environment and (2) using a real robot (the Fetch mobile manipulator) in a table environment. For 4 different iterations the robot is started from different initial locations and the task poses are kept different. For all iterations, task poses are kept outside the robot's reach. As result, the robot had to move its base to an optimal base location from which all task poses could be reached. A simplistic base path planner is incorporated to move the robot base. 

In both simulation and real environments, the task poses were decided based on the point cloud from the robot's depth camera (Refer to Fig.\ref{fig:fetch_task}b and  Fig.\ref{fig:pr2}b for task poses.) The optimal base location and the reachability map, along with manipulator joint solutions for individual task poses, are shown in Fig.\ref{fig:fetch_task}c and  Fig.\ref{fig:pr2}c. The final condition, where the robot successfully reaches a task pose is shown in Fig.\ref{fig:fetch_task}d and  Fig.\ref{fig:pr2}d. Table \ref{tab:real_exp} represents the results from the real world and simulation experiments. In some cases, the robot could not reach the task pose due to failure in motion planning. Since we did not consider environmental obstacle when placing bases, in one scenerio, the robot failed to reach 2 of 3 task poses due to collision. Also, since task poses are defined by depth sensors, depth sensor errors had a substantial negative impact on base placement.

\section{CONCLUSIONS} \label{conclusions}

In this paper, several methods were proposed to find optimal manipulator base locations and robot base locations. The characteristics that distinguishes our process from other available base placement and reachability map creation tool is the time-efficiency, generalizability and user-friendliness. Further, the base solution is not limited to a single solution; the numbers of solutions depend on the tasks and user intent. From Table \ref{tab:basePlacePerformance} we can infer that the robot base placement solution presents significantly improved results vis-a-vis human intuition and other methods. It is not possible for a human being to consider the most optimal base locations by intuition. The limitation of this approach is the input system of the task poses and its exclusion of collision when planning for base placement. The 3D depth cloud sensors are noisy and can provide with incorrect estimation of the environment. Collision in the base placement planning is vital, as the output base pose may be in collision with manipulated workpiece, worktable and other surroundings. Also base placement does not depend only on the reachability of the task poses, the power cost or minimum joint motions should also be considered. The work presented here is available as a self-contained C++ library at http://wiki.ros.org/reuleaux.

\section*{ACKNOWLEDGMENTS}

The authors acknowledge the opportunity provided by ROS-Industrial and Google Summer of Code to accomplish the work presented in this paper. The authors are grateful of Dr.Alba Perez Gracia and Debashree Sheet Makhal for providing encouragement and support throughout the process. We are also thankful to Prof. Maya Cakmak for providing us with a fetch robot we could use for the experiments.

\bibliographystyle{ieeetr}
\bibliography{reu}

\end{document}